\documentclass[runningheads]{llncs}
\usepackage{float}
\usepackage{amsfonts}
\usepackage{graphicx}
\usepackage{booktabs} 
\usepackage[ruled,linesnumbered]{algorithm2e}
\usepackage{subcaption}
\input{epsf}
\begin{document}
\title{Mitigation of Policy Manipulation Attacks on Deep Q-Networks with Parameter-Space Noise 
 }

\titlerunning{Mitigating Policy Manipulation Attacks with Parameter Noise}
%
\author{Vahid Behzadan \and 
Arslan Munir} 

\authorrunning{V. Behzadan et al.}
%
\institute{Kansas State University, Manhattan, KS 66506, USA\\
\email{\{behzadan,amunir\}@ksu.edu}\\
\url{http://blogs.k-state.edu/aisecurityresearch/} \vspace{-2 mm}}
\maketitle              
\begin{abstract}
Recent developments have established the vulnerability of deep reinforcement learning to policy manipulation attacks via intentionally perturbed inputs, known as adversarial examples. In this work, we propose a technique for mitigation of such attacks based on addition of noise to the parameter space of deep reinforcement learners during training. We experimentally verify the effect of parameter-space noise in reducing the transferability of adversarial examples, and demonstrate the promising performance of this technique in mitigating the impact of whitebox and blackbox attacks at both test and training times.

\end{abstract}
Recent years has been the scene to growing interest and advances in deep Reinforcement Learning (RL). By exploiting the superior feature extraction and processing capabilities of deep neural networks, deep RL enables the learning of direct mappings from raw observations of the environment to actions. This enhancement enables the application of classic RL approaches to high-dimensional and complex planning problems, and is shown to achieve human-level or superhuman performance in various cases such as learning to playing the game of Go \cite{silver2016alphago}, playing Atari games \cite{mnih2015human}, robotic manipulation \cite{gu2017deep}, and autonomous navigation of aerial \cite{zhang2016learning} and ground \cite{zhu2017target} vehicles. While the interest in deep RL solutions is extending into numerous domains such as intelligent transportation systems \cite{atallah2017next}, finance \cite{deng2017deep} and critical infrastructure \cite{mohammadi2017semi}, ensuring the security and reliability of such solutions in adversarial conditions is only at its preliminary stages. Recently, Behzadan and Munir \cite{behzadan2017} reported the vulnerability of deep reinforcement learning algorithms to both test-time and training-time attacks using adversarial examples \cite{goodfellow2014explaining}. This work was followed by a number of further investigations (e.g., \cite{huang2017adversarial}, \cite{kos2017delving}), verifying the fragility of deep RL agents to such attacks. Currently, only a few reports (e.g., \cite{behzadan2017whatever}, \cite{lin2017detecting}, \cite{pattanaik2017robust}) concentrate on mitigation and countermeasures, which are mostly focused on approaches based on adversarial training and prediction. 

In this work, we aim to further the research on countering attacks on deep RL by proposing a potential mitigation technique based on employing parameter-space noise exploration during the training of deep RL agents. Recent reports in \cite{plappert2017parameter} and \cite{fortunato2017noisy} demonstrate that addition of adaptive noise to the parameters of deep RL architectures greatly enhances the exploration behavior and convergence speed of such algorithms. Contrary to classical exploration heuristics such as $\epsilon$-greedy \cite{sutton1998reinforcement}, parameter-space noise is iteratively and adaptively applied to the parameters of the learning model, such as weights of the neural network. Accordingly, we hypothesize that the randomness introduced via parameter noise, not only enhances the discovery of more creative and robust policies, but also reduces the effect of whitebox and blackbox adversarial example attacks at both test-time and training-time. 

To this end, we evaluate the performance of Deep Q-Network (DQN) models trained with parameter noise, against the test-time and training-time adversarial example attacks introduced in \cite{behzadan2017}. Main contributions of this work are:
\begin{enumerate}
	\item Proposal of parameter-space noise exploration as a mitigations technique against policy manipulation attacks at both test-time and training-time,
	\item Development of an open-source platform for experimenting with adversarial example attacks on deep RL agents,
	\item Experimental analysis of parameter-space noise for mitigation of test-time whitebox and blackbox attacks on DQN,
	\item Experimental analysis of parameter-space noise for mitigation of training-time policy induction attacks on DQN. 
\end{enumerate}

The remainder of this paper is organized as follows: Section \ref{background} reviews the relevant background of DQN, parameter noise training via the NoisyNet approach, and adversarial examples. Section \ref {Model} describes the attack model adopted in this study. Section \ref{Setup} details the experiment setup, and presents the corresponding results. Section \ref{Conclusion} concludes the paper with remarks on the obtained results.

\section{Background}\label{background}
In this section, we present an overview of the fundamental concepts, upon which this work is based. It must be noted that this overview is not meant to be comprehensive, and thus the interested readers may refer to the suggested references for further details.
\subsection{RL and Deep Q-Networks}

The generic RL problem can be formally modeled as a Markov Decision Process (MDP), described by the tuple $MDP = (S, A, P, R)$, where $S$ is the set of reachable states in the process, $A$ is the set of available actions, $R$ is the mapping of transitions to the immediate reward, and $P$ represents the transition probabilities. At any given time-step $t$, the MDP is at a state $s_t\in S$. The RL agent's choice of action at time $t$, $a_t \in A$ causes a transition from $s_t$ to a state $s_{t+1}$ according to the transition probability $P_{s_t , s_{t+1}}^{a_t}$. The agent receives a reward $r_t = R(s_t, a_t) \in \mathbb{R}$, where $\mathbb{R}$ denotes the set of real numbers, for choosing the action $a_t$ at state $s_t$.

Interactions of the agent with MDP are determined by the policy $\pi$. When such interactions are deterministic, the policy $\pi: S\rightarrow A$ is a mapping between the states and their corresponding actions. A stochastic policy $\pi(s)$ represents the probability of optimality for implementing any action $a\in A$ at state $s$.

The objective of RL is to find the optimal policy $\pi^\ast$ that maximizes the cumulative reward over time at time $t$, denoted by the return function $\hat{R}_t = \sum_{k = 0}^{\infty} \gamma^{k} r_{t+k}$, where $\gamma \in [0,1]$ is the discount factor representing the diminishing worth of rewards obtained further in time, hence ensuring that $\hat{R}$ is bounded.

One approach to this problem is to estimate the optimal value of each action, defined as the expected sum of future rewards when taking that action and following the optimal policy thereafter. The value of an action $a$ in a state $s$ is given by the action-value function $Q$ defined as:
\begin{eqnarray} \label{bellman}
Q(s,a) = R(s, a) + \gamma max_{a'}(Q(s',a'))
\end{eqnarray}

Where $s'$ is the state that emerges as a result of action $a$, and $a'$ is a possible action in state $s'$. The optimal $Q$ value given a policy $\pi$ is hence defined as: $Q^\ast (s, a) = max_{\pi} Q^{\pi} (s, a)$, and the optimal policy is given by $\pi^\ast(s) = \arg\max_a Q(s,a)$

The Q-learning method estimates the optimal action policies by using the Bellman equation $Q_{i+1} (s,a) = \mathbf{E}[R + \gamma \max_a Q_i]$ as the iterative update of a value iteration technique. Practical implementation of Q-learning is commonly based on function approximation of the parametrized Q-function $Q(s,a; \theta) \approx Q^\ast (s,a)$. A common technique for approximating the parametrized non-linear Q-function is via neural network models whose weights correspond to the parameter vector $\theta$. Such neural networks, commonly referred to as Q-networks, are trained such that at every iteration $i$, the following loss function is minimized:
\begin{eqnarray}
L_i(\theta_i) = \mathbf{E}_{s, a\sim \rho(.)} [(y_i - Q(s,a,;\theta_i))^2]
\end{eqnarray}

where $y_i = \mathbf{E}[R + \gamma \max_{a'}Q(s',a';\theta_{i-1}) | s,a]$, and $\rho(s,a)$ is a probability distribution over states $s$ and actions $a$. This optimization problem is typically solved using computationally efficient techniques such as Stochastic Gradient Descent (SGD) \cite{baird1999gradient}.
%
%
%
%
%
%
%


%
%
%
%

Classical Q-networks introduce a number of major problems in the Q-learning process. First, the sequential processing of consecutive observations breaks the \emph{iid} (Independent and Identically Distributed) requirement of training data as successive samples are correlated. Furthermore, slight changes to Q-values leads to rapid changes in the policy estimated by Q-network, thus enabling policy oscillations. Also, since the scale of rewards and Q-values are unknown, the gradients of Q-networks can be sufficiently large to render the backpropagation process unstable.

A Deep Q-network (DQN) \cite{mnih2015human} is a training algorithm designed to resolve these problems. To overcome the issue of correlation between consecutive observations, DQN employs a technique called \emph{experience replay}: instead of training on successive observations, experience replay samples a random batch of previous observations stored in the replay memory to train on. As a result, the correlation between successive training samples is broken and the \emph{iid} setting is re-established. In order to avoid oscillations, DQN fixes the parameters of a network $\hat{Q}$, which represents the optimization target $y_i$. These parameters are then updated at regular intervals by adopting the current weights of the Q-network. The issue of unstability in backpropagation is also solved in DQN by normalizing the reward values to the range $[-1,+1]$, thus preventing Q-values from becoming too large.

Mnih et al. \cite{mnih2015human} demonstrate the application of this new Q-network technique to end-to-end learning of Q values in playing Atari games based on observations of pixel values in the game environtment. To capture the movements in the game environment, Mnih et al. use stacks of 4 consecutive image frames as the input to the network. To train the network, a random batch is sampled from the previous observation tuples $(s_t, a_t, r_t, s_{t+1})$, where $r_t$ denotes the reward at time $t$. Each observation is then processed by 2 layers of convolutional neural networks to learn the features of input images, which are then employed by feed-forward layers to approximate the Q-function. The target network $\hat{Q}$, with parameters $\theta^{-}$, is synchronized with the parameters of the original $Q$ network at fixed periods intervals. i.e., at every $i$th iteration,  $\theta^-_{t} = \theta_t$, and is kept fixed until the next synchronization. The target value for optimization of DQN thus becomes:

\begin{eqnarray}
y'_t \equiv r_{t+1} + \gamma max_{a'} \hat{Q}(S_{t+1}, a'; \theta^-)
\end{eqnarray}

Accordingly, the training process can be stated as:

\begin{eqnarray}\label{SGD}
min_{a_t} (y'_t - Q(s_t, a_t, \theta))^2
\end{eqnarray}

As for the exploration mechanism, the original DQN employs $\epsilon$-greedy, which  monotonically decreases the probability of taking random actions as the training progresses \cite{sutton1998reinforcement}. 

\subsection{NoisyNets}
Introduced by Fortunato et al. \cite{fortunato2017noisy}, NoisyNet is a type of neural network whose biases and weights are iteratively perturbed during training by a parametric function of the noise. Such a neural network can be represented by $y = f_\theta (x)$, parametrized by the vector of noisy parameters $\theta = \mu +\Sigma \ast \epsilon$, where $\tau = (\mu, \Sigma)$ is a set of vectors representing learnable parameters, $\epsilon$ is a vector of zero-mean noise with fixed statistics, and $\ast$ is element-wise multiplication. In \cite{fortunato2017noisy}, the modified DQN algorithm is proposed as follows: first, $\epsilon$-greedy is omitted, and instead the value function is greedily optimized. Second, the fully connected layers of the value function are parametrized as a NoisyNet, whose parameter values are drawn from a noisy parameter distribution after every replay step. The noise distribution used in \cite{fortunato2017noisy} is factorized Gaussian noise. During replay, the current NoisyNet parameter samples are held constant, while at the optimization of each action step, the parameters are re-sampled. The parametrized action-value function $Q(x,a,\epsilon; \tau)$ can be treated as a random variable, and is employed accordingly in the optimization function. Further details of this approach and a similar proposal can be found in \cite{fortunato2017noisy} and \cite{plappert2017parameter}, respectively.

\subsection{Adversarial Examples}

In \cite{szegedy2013intriguing}, Szegedy et al. report an intriguing discovery: several machine learning models, including deep neural networks, are vulnerable to adversarial examples. That is, these machine learning models misclassify inputs that are only slightly different from correctly classified samples drawn from the data distribution. Furthermore, it was found \cite{papernot2016limitations} that a wide variety of models with different architectures trained on different subsets of the training data misclassify the same adversarial example.

This suggests that adversarial examples expose fundamental blind spots in machine learning algorithms. The issue can be stated as follows: Consider a machine learning system $M$ and a benign input sample $C$ which is correctly classified by the machine learning system, i.e. $M(C) = y_{true}$. According to the report of Szegedy \cite{szegedy2013intriguing} and many proceeding studies \cite{papernot2016limitations}, it is possible to construct an adversarial example $A = C + \delta$, which is perceptually indistinguishable from $C$, but is classified incorrectly, i.e. $M(A) \neq y_{true}$.

Adversarial examples are misclassified far more often than examples that have been perturbed by random noise, even if the magnitude of the noise is much larger than the magnitude of the adversarial perturbation \cite{goodfellow2014explaining}. According to the objective of adversaries, adversarial example attacks are generally classified into the following two categories: 

\begin{enumerate}
	
	\item Misclassification attacks, which aim for generating examples that are classified incorrectly by the target network
	
	\item Targeted attacks, whose goal is to generate samples that the target misclassifies into an arbitrary class designated by the attacker.
	
\end{enumerate}

To generate such adversarial examples, several algorithms have been proposed, such as the Fast Gradient Sign Method (FGSM) by Goodfellow et al., \cite{goodfellow2014explaining}, and the Jacobian Saliency Map Algorithm (JSMA) approach by Papernot et al., \cite{papernot2016limitations}. A grounding assumption in many of the crafting algorithms is that the attacker has complete knowledge of the target neural networks such as its architecture, weights, and other hyperparameters. In response, Papernot et al. \cite{papernot2016practical} proposed the first blackbox approach to generating adversarial examples. This method exploits the transferability of adversarial examples: an adversarial example generated for a neural network classifier applies to most other neural network classifiers that perform the same classification task, regardless of their architecture, parameters, and even the distribution of training data. Accordingly, the approach of \cite{papernot2016practical} is based on generating a replica of the target network. To train this replica, the attacker creates and trains over a dataset from a mixture of samples obtained by observing target's interaction with the environment, and synthetically generated inputs and label pairs. Once trained, any of the algorithms that require knowledge of the target network for crafting adversarial examples can be applied to the replica. Due to the transferability of adversarial examples, the perturbed data points generated for the replica network can induce misclassifications in many of the other networks that perform the same task.

\section{Attack Model}\label{Model}
We consider an attacker whose goal is to perturb the optimality of actions taken by a DQN agent through either perturbing the observations of the agent the test-time, or inducing an arbitrary policy $\pi_{adv}$ on the target DQN at training time. In whitebox attacks, the attacker has complete knowledge of the target. On the other hand, a blackbox attacker has no knowledge of the target's exact architecture and parameters, but is assumed to be capable of estimating those based on the conventions applied to the input type (e.g. image and video input may indicate a convolutional neural network, speech and voice data point towards a recurrent neural network, etc.). 

In this model, The attacker is assumed to have minimal \emph{a priori} information of the target's model and parameters, such as the type and format of inputs to the DQN, as well as its reward function $R$ and an estimate for the frequency of updating the $\hat{Q}$ network. Furthermore, the attacker has no direct influence on the target's architecture and parameters, including its reward function, parameter noise, and the optimization mechanism. The only parameter that the attacker can directly manipulate is the configuration of the environment observed by the target. For instance, in the case of DQN agents learning to play Atari games \cite{mnih2015human}, the attacker may change pixel values of the game's frames, but not the score. We assume that the attacker is capable of changing the state before it is observed by the target by predicting future states, through approaches such as having a quicker action speed than the target's sampling rate, or by introducing a delay between generation of the new environment and its observation by the target. 

To avoid detection, we impose an extra constraint on the attack such that the magnitude of perturbations applied in each configuration must be smaller than a constant value denoted by $\lambda$. Also, we do not limit the attacker's domain of perturbations.

As discussed in Section \ref{background}, the DQN framework of Mnih et al. \cite{mnih2015human} can be seen as consisting of two neural networks, one is the native Q-network which performs the image processing and function approximation, and the other is the target network $\hat{Q}$ network whose architecture and parameters are copies of the native network sampled once every $c$ iterations. DQN is trained through optimizing the loss function of equation \ref{SGD} by SGD. Behzadan and Munir \cite{behzadan2017} demonstrated that the function approximators of DQN are also vulnerable to adversarial example attacks. In other words, the set of all possible inputs to the approximated function $\hat{Q}$ contains elements which cause the approximated functions to generate outputs that are different from the output of the original $Q$ function. 

Consequently, the attacker can manipulate the learning process of DQN by crafting states $s_t$ such that $\hat{Q}(s_{t+1}, a; \theta^-_{t})$ identifies an incorrect choice of optimal action at $s_{t+1}$. If the attacker is capable of crafting adversarial inputs $s'_t$ and $s'_{t+1}$ such that the value of Equation \ref{SGD} is minimized for a specific action $a'$, then the policy learned by DQN at this time-step is optimized towards suggesting $a'$ as the optimal action given the state $s_t$. At every time step of training this replica, the attacker observes interactions of its target with the environment $(s_t, a_t, r_t, s_{t+1})$. If the resulting state is not terminal, the attacker then calculates the perturbation vectors $\hat{\delta}_{t+1}$ for the next state $s_{t+1}$ such that $max_{a'} \hat{Q}(s_{t+1} + \hat{\delta}_{t+1}, a'; \theta^-_{t})$ causes $\hat{Q}$ to generate its maximum when $a' = {\pi^\ast_{adv}}(s_{t+1})$, i.e., the maximum reward at the next state is obtained when the optimal action taken at that state is determined by the attacker's policy. The attacker then reveals the perturbed state $s_{t+1}$ to the target, and re-trains the replica based on the new state and action.

%
%

This is procedurally similar to targeted misclassification attacks described in Section \ref{background}, which aim to find minimal perturbations to an input sample such that the classifier assigns the maximum value of likelihood to an incorrect target class. Therefore, the adversarial example crafting techniques developed for classifiers such as FGSM can be employed to obtain the perturbation vector $\hat{\delta}_{t+1}$. 

\begin{figure}[!h]
	
	\centering
	
	\includegraphics[width = 4in]{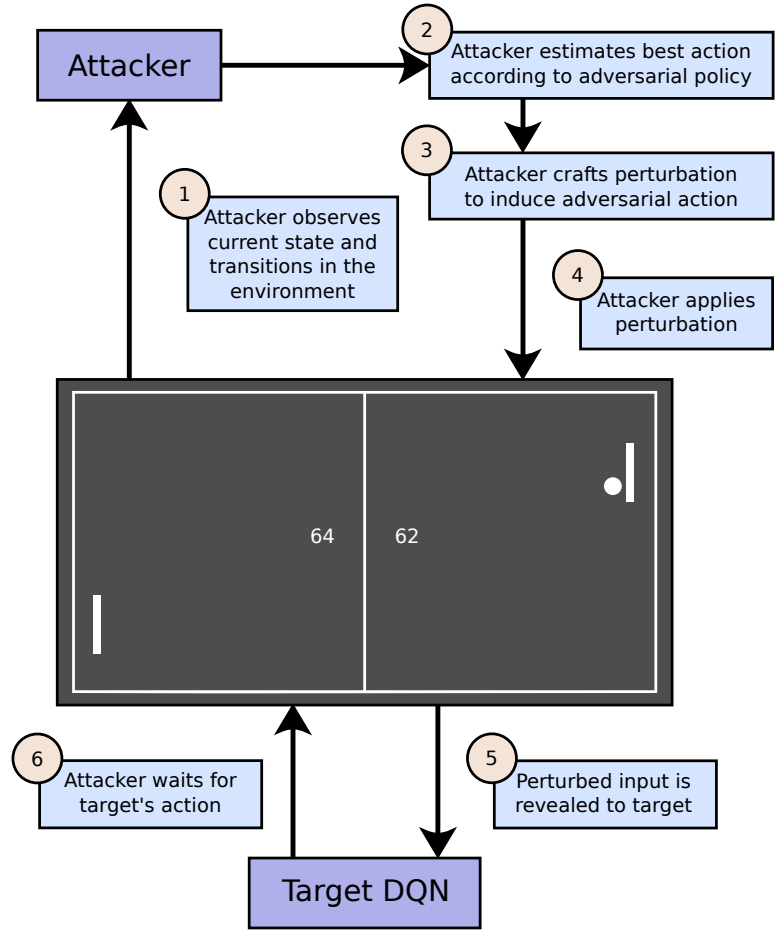}
	
	\caption{Exploitation cycle of policy induction attack}
	
	\label{fig:exploit}
	
\end{figure}

Accordingly, Behzadan and Munir \cite{behzadan2017} divide this attack into the two phases of initialization and exploitation. The initialization phase implements processes that must be performed before the target begins interacting with the environment, which are:

\begin{enumerate}
	\item Train a DQN based on attacker's reward function $R'$ to obtain the adversarial policy $\pi^\ast_{adv}$
	\item Create a replica of the target's DQN and initialize with random parameters 
\end{enumerate}

The exploitation phase implements the attack process and crafting adversarial inputs, such that the target DQN performs an action dictated by $\pi^\ast_{adv}$. This phase constitutes an attack cycle depicted in figure \ref{fig:exploit}. The cycle initiates with the attacker's first observation of the environment, and runs in tandem with the target's operation. 

\section{Experimental Verification} \label{Setup}
To evaluate the effectiveness of NoisyNet in mitigation of adversarial example attacks, we study the performance of this architecture in comparison to the original DQN setup. Following the standard benchmarks of DQN, our experimental environments consist of 3 Atari 2600 games, namely Enduro, Assault, and Blackout. We train 4 models for each environment, 2 models based on the original DQN and $\epsilon$-greedy exploration, and 2 models based on the NoisyNet architecture. The neural network configuration of both models follows that of the original DQN proposal by Mnih et al. \cite{mnih2015human}, while the parameter noise configuration is based on the setup presented in \cite{fortunato2017noisy}.

We implemented the experimentation platform in TensorFlow using OpenAI Gym \cite{brockman2016openai} for emulating the game environment and Cleverhans \cite{papernot2016cleverhans} for crafting the adversarial examples. Our DQN implementation is a modified version of the module in OpenAI Baselines \cite{baselines}, while the NoisyNet implementation is based on the algorithm described in \cite{fortunato2017noisy}. We have published our platform at \cite{RLAttack} for open-source use in further research in this area.

For the purposes of this study, we consider FGSM for crafting adversarial examples, with the perturbation limit $\lambda = 1.0/255.0$. Similar to the work in \cite{kos2017delving}, the initiation of attacks occurs after the learned Q-function begins converging towards the optimal value. 
\vspace{-12mm}
\subsection{Test-time Attacks} \label{Results}
\vspace{-2mm}
Parameter noise training in NoisyNet is expected to enhance the exploration criteria of the agent and hence facilitate learning more creative and accurate policies. Accordingly, we hypothesize that the action-value function learned in NoisyNet is better generalized than the original, and can be more resilient to non-targeted adversarial example attacks at test-time. Similarly, the addition of random noise to the parameters of NoisyNet can potentially impede the transferability of adversarial examples, and hence enhance the resilience of NoisyNet to blackbox attacks. To test this hypothesis, we compare the performance of NoisyNet and DQN models to whitebox and blackbox attacks after $2e8$ iterations of training. 

Figure \ref{Test-time} presents the results of this experiment. It is observed that in all three environments, the impact of adversarial example perturbation in the performance of NoisyNet is less severe than that of the original DQN, thereby verifying our general hypothesis. Furthermore, comparison of performance under blackbox attacks demonstrates significant improvements in Noisynets, as depicted in all three cases. A preliminary interpretation of this observation is that the randomization of model parameters reduces the transferability of adversarial examples generated for a replicated model.

\begin{figure}[h]
	\begin{subfigure}[h]{0.45\textwidth}
		\includegraphics[width=\textwidth]{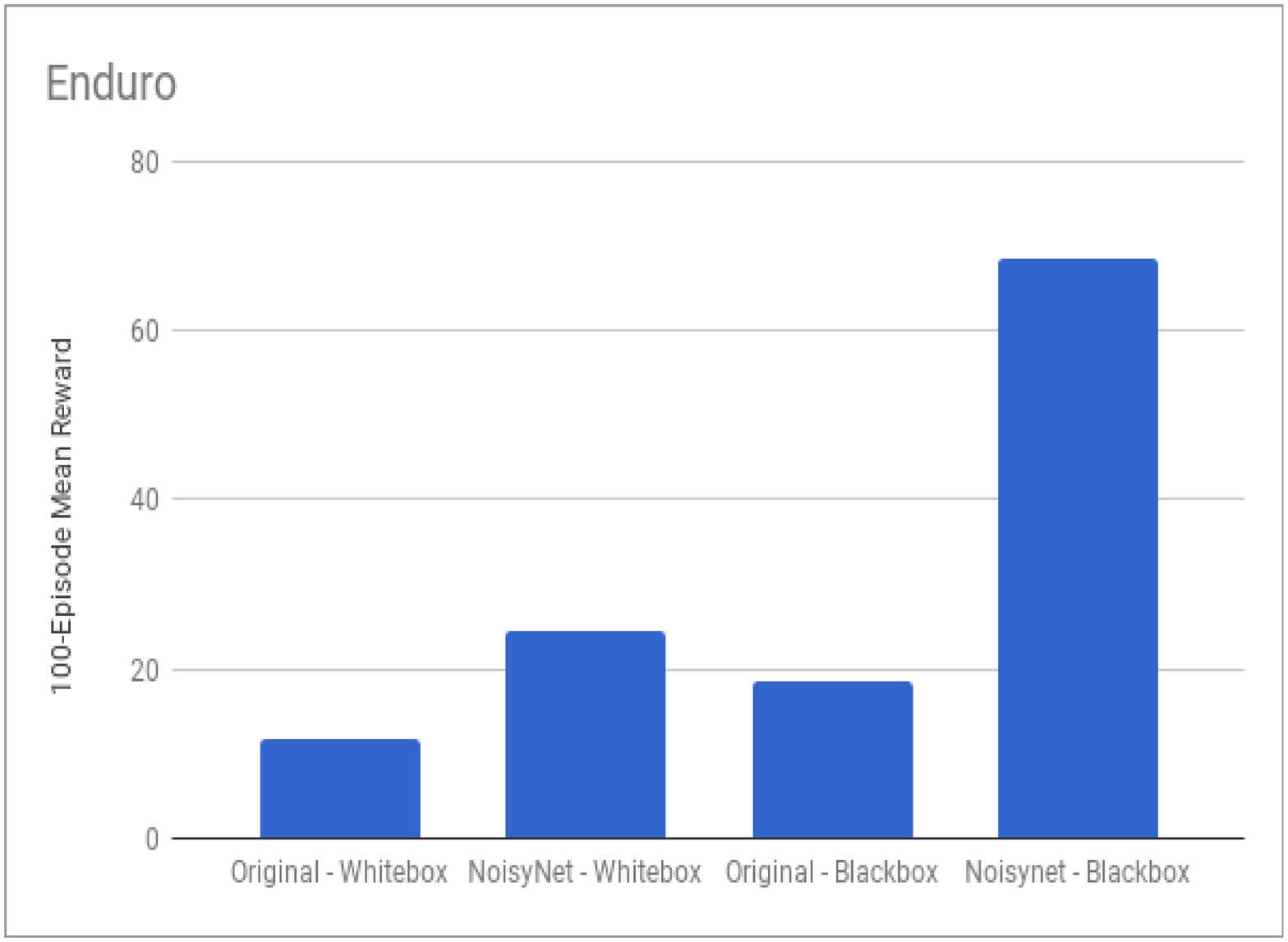}
	\end{subfigure}
	\begin{subfigure}[h]{0.45\textwidth}
		\includegraphics[width=\textwidth]{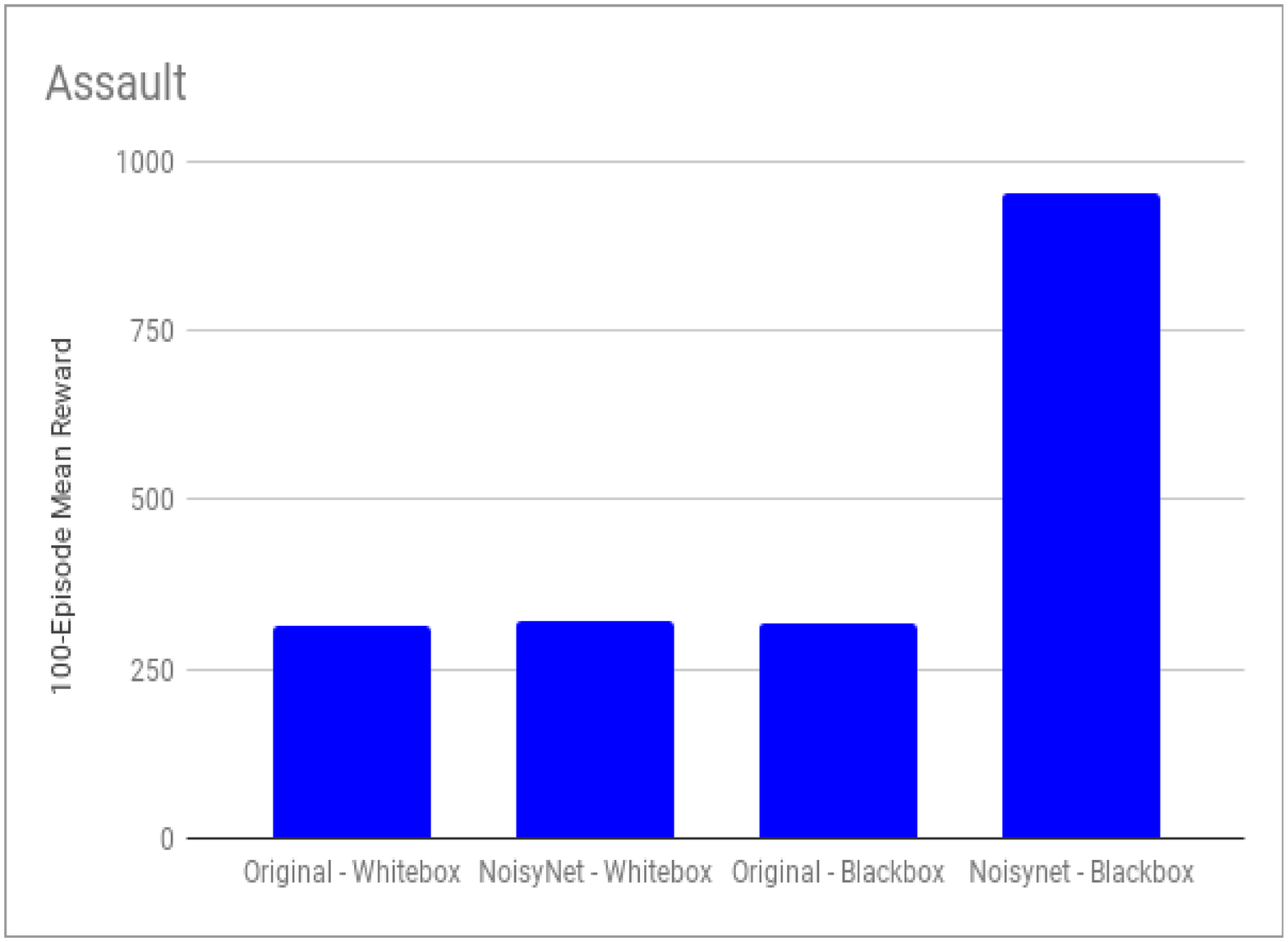}
	\end{subfigure}
	\begin{subfigure}[h]{0.45\textwidth}
		\includegraphics[width=\textwidth]{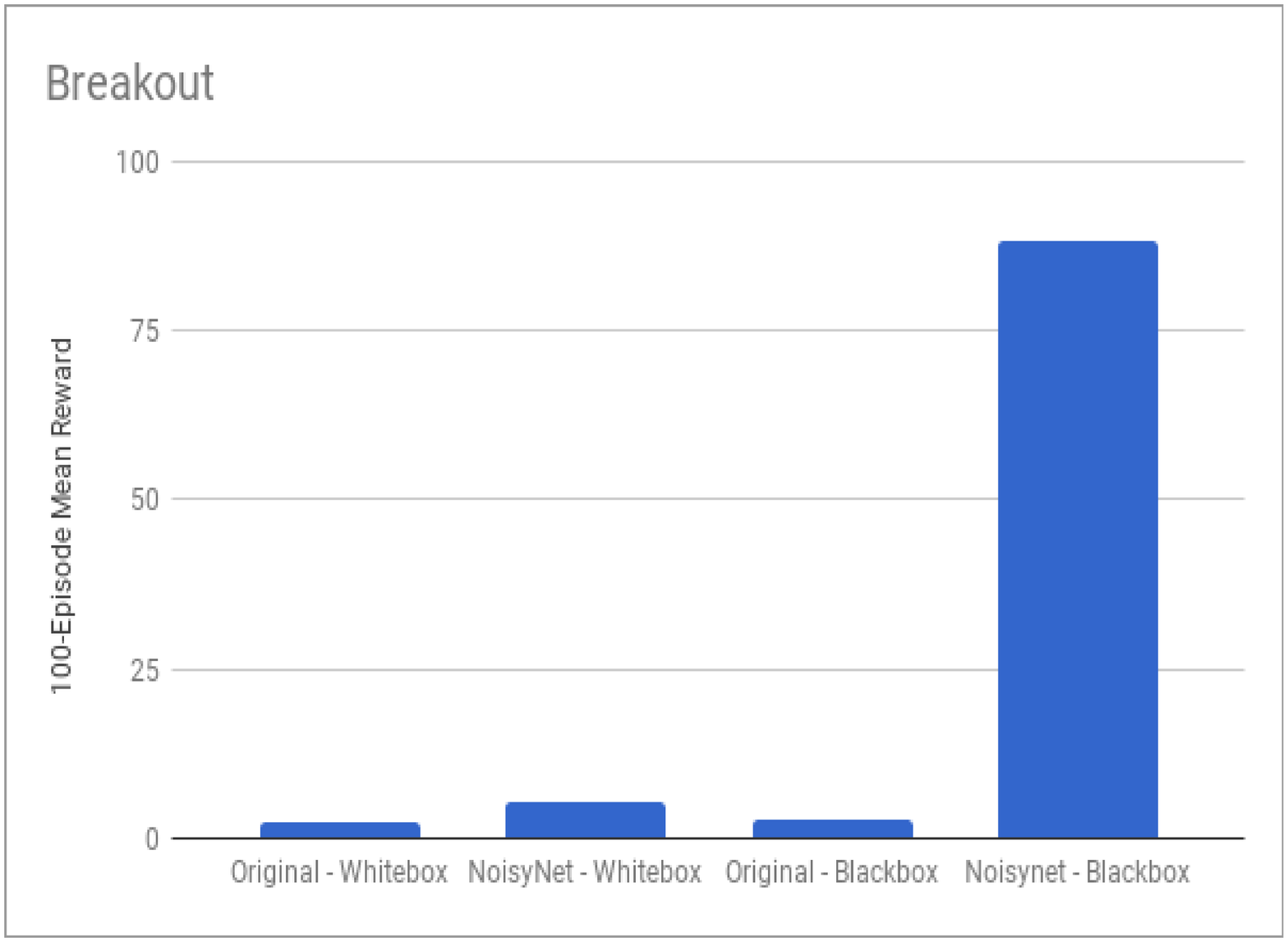}
	\end{subfigure}
	\caption{Comparison of whitebox and blackbox attacks at test-time} 
\end{figure} \label{Test-time}

\subsection{Training-time Attacks}
In \cite{behzadan2017} and \cite{kos2017delving}, the impact of training-time adversarial example attacks on the policy learning is demonstrated. Similar to the case of test-time attacks, we hypothesize that the reduced transferability and enhanced generalization of NoisyNet can potentially provide greater resilience to blackbox adversarial example attacks during training. To this end, we investigated the performance of NoisyNet and DQN to the training-time attack methodology described in Section \ref{Model} \cite{behzadan2017}. 
\begin{figure}[H]
	\begin{subfigure}[h]{0.45\textwidth}
		\includegraphics[width=\textwidth]{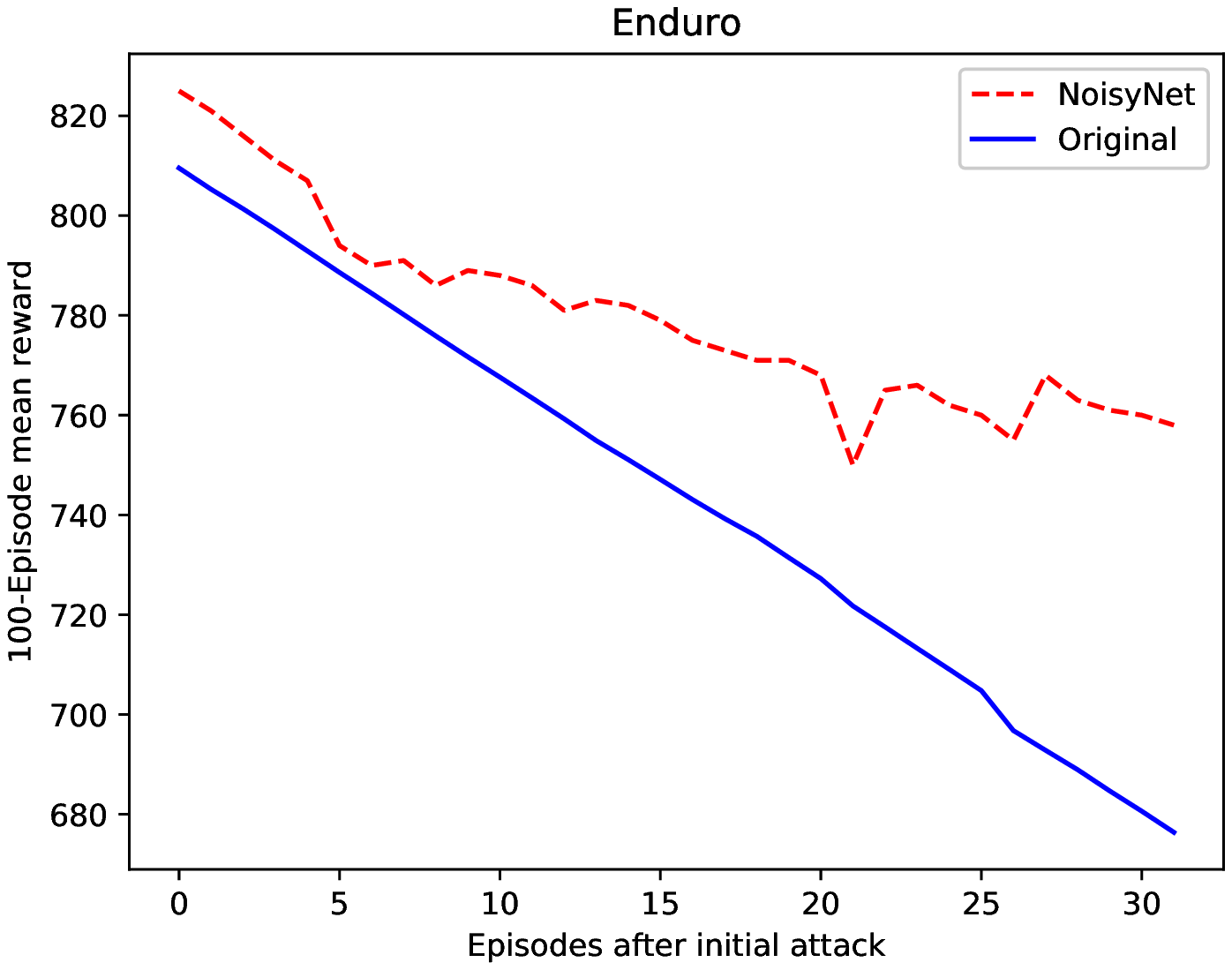}
	\end{subfigure}
	\begin{subfigure}[h]{0.45\textwidth}
		\includegraphics[width=\textwidth]{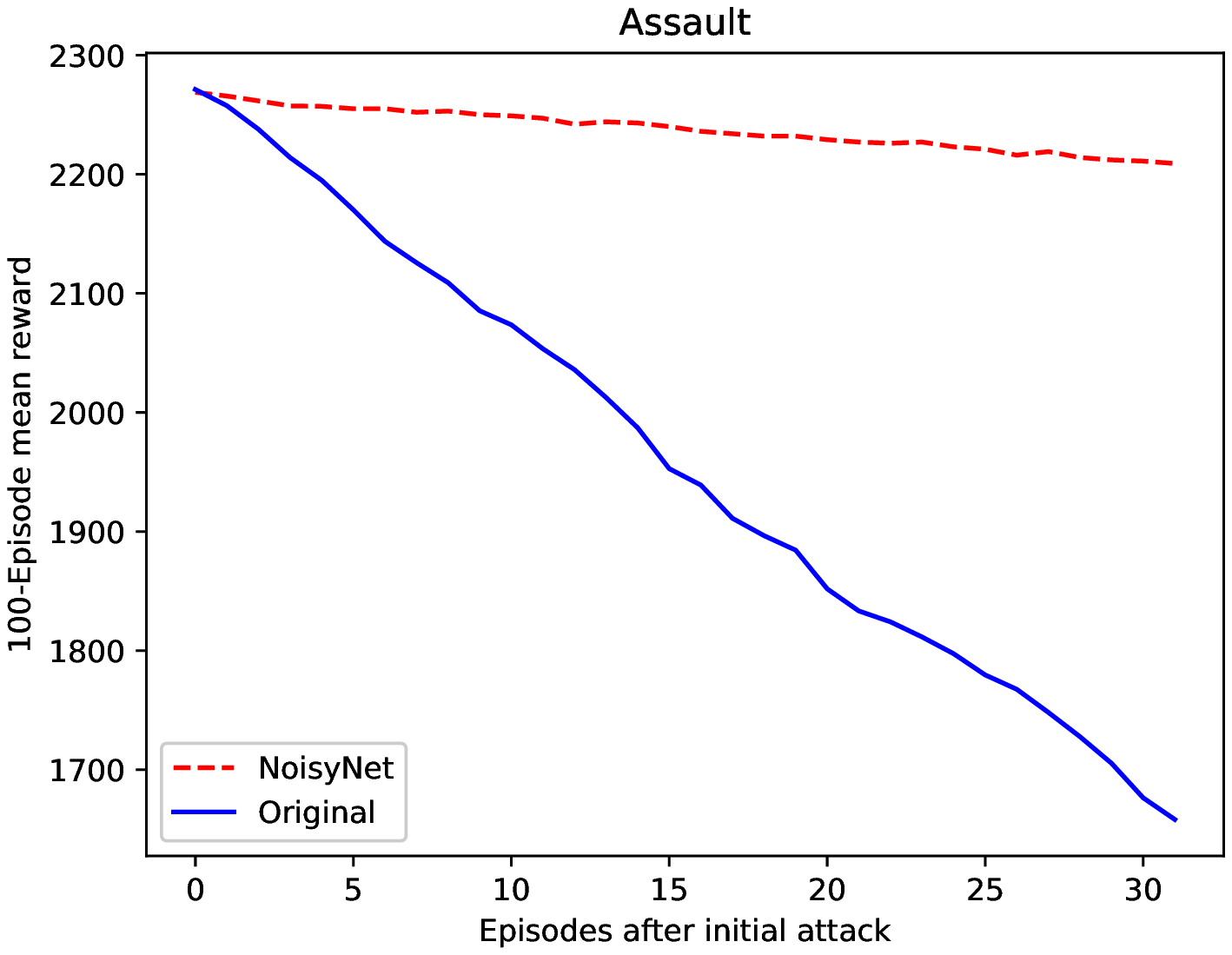}
	\end{subfigure}
	\begin{subfigure}[h]{0.45\textwidth}
		\includegraphics[width=\textwidth]{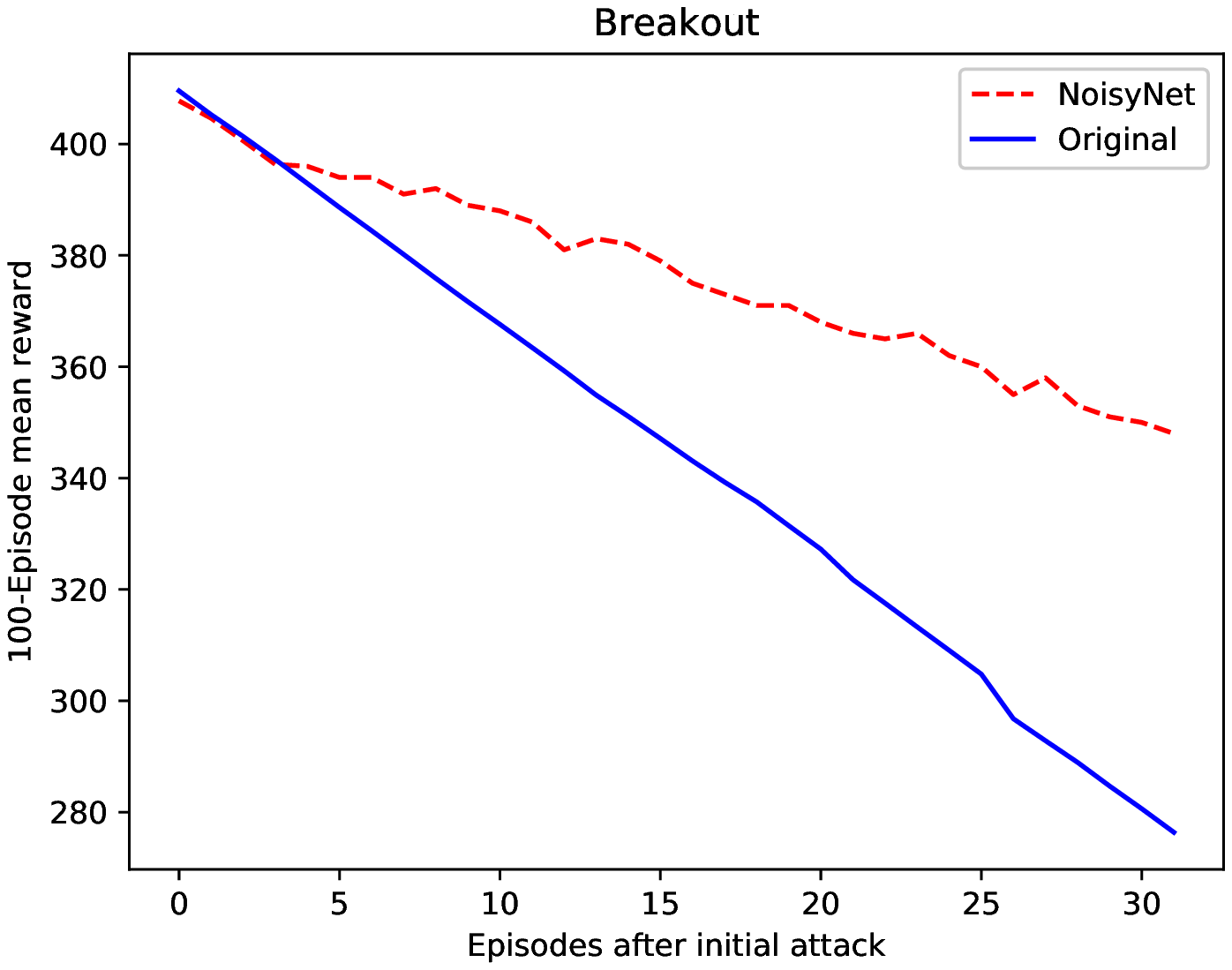}
	\end{subfigure}
	\caption{Comparison of blackbox attacks at training-time} 
\end{figure} \label{Training-Time}
Figure \ref{Training-Time} presents the results of this experiment. It can be seen that in all three environments, performance of the original DQN consistently deteriorates under training-time attacks, as reported in \cite{behzadan2017} and \cite{kos2017delving}. On the other hand, while the performance of NoisyNet is also subject to deterioration, it demonstrates significantly stronger resilience to this attack, and in the case of Assault remains almost unaffected by adversarial perturbations. These results verify the original hypothesis, and hence the efficacy of parameter noise in mitigating the impact of training-time attacks.

\section{Conclusion} \label{Conclusion}
Through experimental analysis, we investigated the effect of parameter noise in mitigation of adversarial example attacks on Deep Q-Networks (DQN). Considering the reported enhancing effect of parameter noise in reinforcement learning and exploration, as well as the inherent randomization of such techniques, we demonstrated that compared to the original DQN, noisy DQN architectures provide better resilience to adversarial perturbations at test-time, and reduce susceptibility to transferability of adversarial examples. Furthermore, we demonstrate that noisy DQN is significantly more resilient to blackbox attacks at training-time, and learn in a greatly more robust manner in comparison to plain DQN architectures. These results present a promising starting point for further experimental and analytical analysis of employing parameter-space noise exploration for enhancement of resilience and robustness in deep reinforcement learning.

\vspace{-3 mm}
\end{document}